\documentclass[10pt,twocolumn,letterpaper]{article}

\usepackage{cvpr}              
\usepackage[dvipsnames]{xcolor}

\definecolor{cvprblue}{rgb}{0.21,0.49,0.74}
\usepackage[pagebackref,breaklinks,colorlinks,citecolor=cvprblue]{hyperref}
\usepackage{amsmath,amssymb,amsthm}

\usepackage{multirow}
\usepackage{colortbl}
\definecolor{mylightblue}{rgb}{0.85, 0.9, 0.95} 
\usepackage{makecell}
\usepackage{array}
\newcolumntype{M}[1]{>{\centering\arraybackslash}m{#1}}
\usepackage{mathtools,tikz}
\DeclareRobustCommand\sampleline[1]{%
  \tikz\draw[#1] (0,0) (0,\the\dimexpr\fontdimen22\textfont2\relax)
  -- (1.5em,\the\dimexpr\fontdimen22\textfont2\relax);%
}

\makeatletter
 \def\hlinewd#1{%
      \noalign{\ifnum0=`}\fi\hrule \@height #1 \futurelet
      \reserved@a\@xhline}

\newcommand\blfootnote[1]{%
  \begingroup
  \renewcommand\thefootnote{}\footnote{#1}%
  \addtocounter{footnote}{-1}%
  \endgroup
}

\title{Diffusion-driven GAN Inversion for Multi-Modal Face Image Generation}

\author{
Jihyun Kim\textsuperscript{\rm 1,2},
Changjae Oh\textsuperscript{\rm 3},
Hoseok Do\textsuperscript{\rm 2},
Soohyun Kim\textsuperscript{\rm 2},
Kwanghoon Sohn\textsuperscript{\rm 1,4}\thanks{Corresponding author} \\
\textsuperscript{\rm 1} Yonsei University
\textsuperscript{\rm 2} AI Lab, CTO Division, LG Electronics\\
\textsuperscript{\rm 3} Queen Mary University of London
\textsuperscript{\rm 4} Korea Institute of Science and Technology (KIST)\\
{\tt\small \{hyunys21, khsohn\}@yonsei.ac.kr c.oh@qmul.ac.uk \{hoseok.do, soohyun1.kim\}@lge.com
}
}

\begin{document}
\maketitle  

\blfootnote{This research was supported by the National Research Foundation of Korea (NRF) grant funded by the Korea government (MSIP) (NRF2021R1A2C2006703).}
\vspace{-10pt}
\begin{abstract}
We present a new multi-modal face image generation method that converts a text prompt and a visual input, such as a semantic mask or scribble map, into a photo-realistic face image. To do this, we combine the strengths of Generative Adversarial networks (GANs) and diffusion models (DMs) by employing the multi-modal features in the DM into the latent space of the pre-trained GANs. We present a simple mapping and a style modulation network to link two models and convert meaningful representations in feature maps and attention maps into latent codes. With GAN inversion, the estimated latent codes can be used to generate 2D or 3D-aware facial images. We further present a multi-step training strategy that reflects textual and structural representations into the generated image. Our proposed network produces realistic 2D, multi-view, and stylized face images, which align well with inputs. We validate our method by using pre-trained 2D and 3D GANs, and our results outperform existing methods. Our project page is available at \small{\url{https://github.com/1211sh/Diffusion-driven_GAN-Inversion/}}.
\end{abstract}
\vspace{-8pt}
\begin{figure}[!t]
    \centering 
    \includegraphics[width=0.91\linewidth]{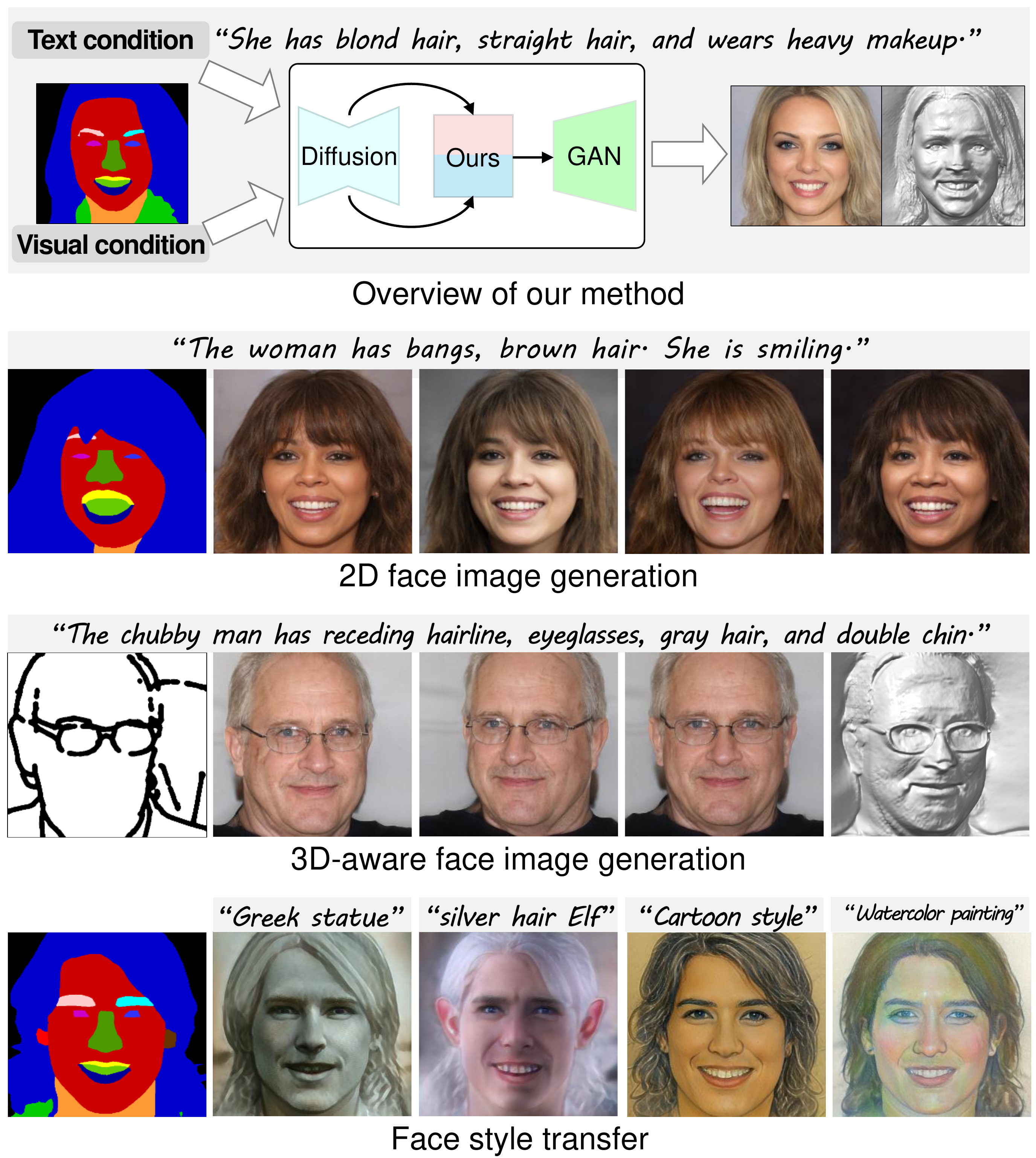}   
    \caption{We present a method to map the diffusion features to the latent space of a pre-trained GAN, which enables diverse tasks in multi-modal face image generation and style transfer. Our method can be applied to 2D and 3D-aware face image generation.
    }
    \vspace{-10pt}
    \label{fig:intro}
\end{figure}
\section{Introduction}
\label{sec:intro}
\vspace{-5pt}
In recent years, multi-modal image generation has achieved remarkable success, driven by the advancements in Generative Adversarial Networks (GANs)~\cite{goodfellow2014generative} and diffusion models (DMs)~\cite{dhariwal2021diffusion,ho2020denoising, song2020score}.
Facial image processing has become a popular application for a variety of tasks, including face image generation~\cite{karras2019style, Patashnik_2021_ICCV}, face editing~\cite{shen2020interpreting,choi2018stargan,pan2023drag,liu2019stgan,nichol2021glide,do2023quantitative}, and style transfer~\cite{chong2022jojogan,zhang2023inversion}.
Many tasks typically utilize the pre-trained StyleGAN~\cite{karras2019style, karras2020analyzing}, which can generate realistic facial images and edit facial attributes by manipulating the latent space using GAN inversion~\cite{Patashnik_2021_ICCV, xia2021tedigan, richardson2021encoding}. In these tasks, using multiple modalities as conditions is becoming a popular approach, which improves the user's controllability in generating realistic face images. However, existing GAN inversion methods~\cite{xia2021tedigan, sun2022ide} have poor alignment with inputs as they neglect the correlation between multi-modal inputs. They struggle to map the different modalities into the latent space of the pre-trained GAN, such as by mixing the latent codes or optimizing the latent code converted from a given image according to the input text.

Recently, DMs have increased attention in multi-modal image generation thanks to the stability of training and the flexibility of using multiple modalities as conditions. DMs~\cite{voynov2023sketch, kawar2023imagic, tumanyan2023plug} can control the multiple modalities and render diverse images by manipulating the latent or attention features across the time steps.
However, existing text-to-image DMs rely on an autoencoder and text encoder, such as CLIP~\cite{radford2021learning}, trained on unstructured datasets collected from the web~\cite{perera2023analyzing, saharia2022photorealistic} that may lead to unrealistic image generation.

Moreover, some approaches address multi-modal face image generation in a 3D domain.
In GAN inversion~\cite{sun2022ide, gal2021stylegannada}, multi-view images can be easily acquired by manipulating the latent code with pre-trained 3D GANs. While DMs are inefficient in learning 3D representation, which has the challenge to generate multi-view images directly due to the lack of 3D ground-truth (GT) data for training~\cite{luo2021diffusion,shue20233d}. They can be used as a tool to acquire training datasets for 3D-aware image generation~\cite{kim2023datid, ma2023free}.

In this paper, we present a versatile face generative model that uses text and visual inputs. We propose an approach that takes the strengths of DMs and GAN and generates photo-realistic images with flexible control over facial attributes, which can be adapted to 2D and 3D domains, as illustrated in Figure~\ref{fig:intro}.
Our method employs a latent mapping strategy that maps the diffusion features into the latent space of a pre-trained GAN using multi-denoising step learning, producing the latent code that encodes the details of text prompts and visual inputs.

In summary, our main contributions are:
\begin{itemize}\setlength{\itemsep}{-0.2mm}
\item[(\romannumeral 1)] We present a novel method to link a pre-trained GAN (StyleGAN~\cite{karras2020analyzing}, EG3D~\cite{chan2022efficient}) and DM (ControlNet~\cite{zhang2023adding}) for multi-modal face image generation.
\item[(\romannumeral 2)] We propose a simple mapping network that links pre-trained GAN and DM’s latent spaces and an attention-based style modulation network that enables the use of meaningful features related to multi-modal inputs.
\item[(\romannumeral 3)] We present a multi-denoising step training strategy that enhances the model's ability to capture the textual and structural details of multi-modal inputs.
\item[(\romannumeral 4)] Our model can be applied for both 2D- and 3D-aware face image generation without additional data or loss terms and outperforms existing DM- and GAN-based methods.
\end{itemize}

\section{Related Work}
\label{sec:related_gan_inversion}
\vspace{-1pt}
\subsection{GAN Inversion}
\vspace{-5pt}
GAN inversion approaches have gained significant popularity in the face image generation task ~\cite{yang2023out,liu2023fine,sun2022ide,chong2022jojogan} using the pre-trained 2D GAN, such as StyleGAN~\cite{karras2019style, karras2020analyzing}. This method has been extended to 3D-aware image generation~\cite{yin20223d, yin2023nerfinvertor, ko20233d} by integrating 3D GANs, such as EG3D~\cite{chan2022efficient}.
GAN inversion can be categorized into learning-based, optimization-based, and hybrid methods.
Optimization-based methods~\cite{saha2021loho, zhu2021barbershop} estimate the latent code by minimizing the difference between an output and an input image. 
Learning-based methods~\cite{alaluf2021restyle, tov2021designing} train an encoder that maps an input image into the latent space of the pre-trained GAN. 
Hybrid methods~\cite{zhu2020domain, xia2021tedigan} combine these two methods, producing an initial latent code and then refining it with additional optimizations.
Our work employs a learning-based GAN inversion, where a DM serves as the encoder.
We produce latent codes by leveraging semantic features in the denoising U-Net, which can generate images with controlled facial attributes. 

\begin{figure*}[!hbt]
\centering
\includegraphics[width=1.0\linewidth]{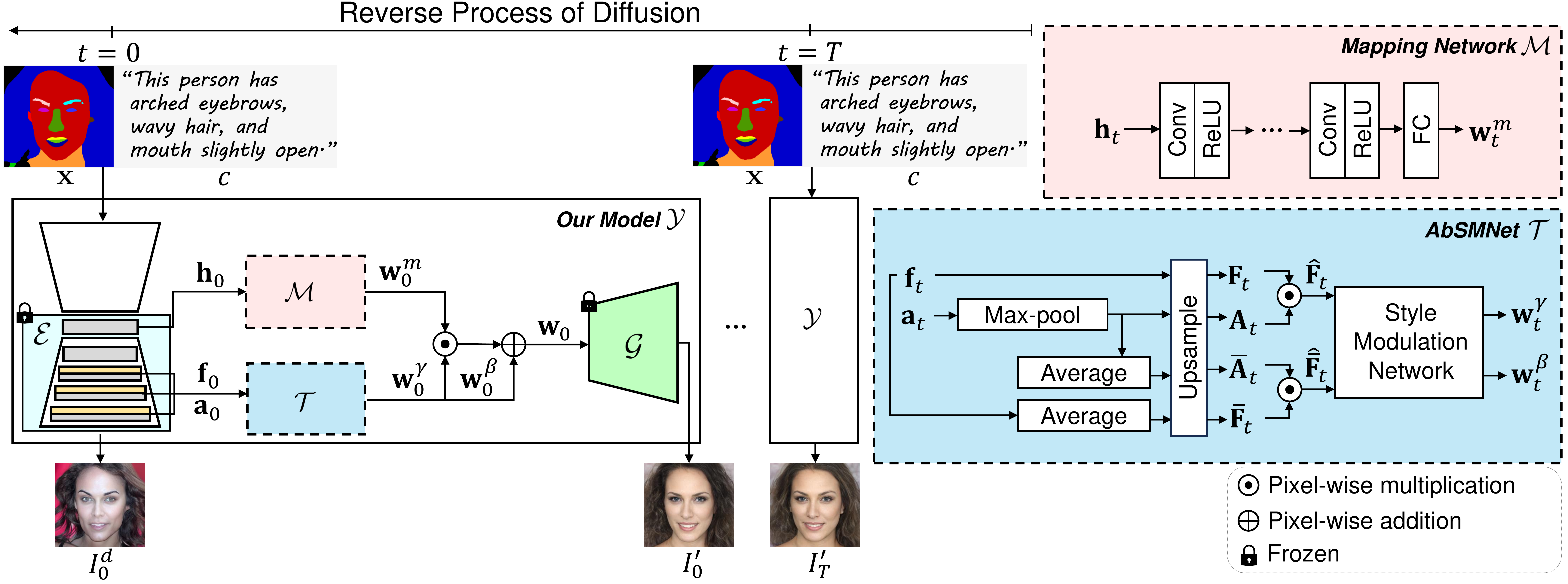}
\caption{Overview of our method. We use a diffusion-based encoder $\mathcal{E}$, the middle and decoder blocks of a denoising U-Net, that extracts the semantic features $\mathbf{h}_t$, intermediate features $\mathbf{f}_t$, and cross-attention maps $\mathbf{a}_t$ at denoising step $t$.
We present the mapping network $\mathcal{M}$ (Sec.~\ref{sec:MappingNet}) and the attention-based style modulation network (AbSMNet) $\mathcal{T}$ (Sec.~\ref{sec:AbSMNet}) that are trained across $t$ (Sec.~\ref{sec:Losses}). $\mathcal{M}$ converts $\mathbf{h}_t$ into the mapped latent code $\mathbf{w}^m_t$, and $\mathcal{T}$ uses $\mathbf{f}_t$ and $\mathbf{a}_t$ to control the facial attributes from the text prompt $c$ and visual input $\mathbf{x}$. The modulation codes $\mathbf{w}^\gamma_t$ and $\mathbf{w}^\beta_t$ are then used to scale and shift $\mathbf{w}^m_t$ to produce the final latent code, $\mathbf{w}_t$, that is fed to the pre-trained GAN $\mathcal{G}$. We obtain the generation output ${I'_t}$ from our model $\mathcal{Y}$ and we use the image $I^d_0$ from the U-Net after the entire denoising process for training $\mathcal{T}$ (Sec.~\ref{sec:Losses}). Note that only the networks with the dashed line (\sampleline{dashed}) are trainable, while others are frozen.
}
\label{fig:overview_model}
\end{figure*}
\subsection{Diffusion Model for Image Generation}
\vspace{-5pt}
Many studies have introduced text-to-image diffusion models~\cite{nichol2021glide, saharia2022photorealistic, rombach2022high} that generate images by encoding multi-modal inputs, such as text and image, into latent features via foundation models~\cite{radford2021learning} and mapping them to the features of denoising U-Net via an attention mechanism. ControlNet~\cite{zhang2023adding} performs image generation by incorporating various visual conditions ($\eg$, semantic mask, scribbles, edges) and text prompts. Image editing models using DMs~\cite{meng2021sdedit, hertz2022prompt, kim2023dense, kwon2022diffusion, jeong2023training} have exhibited excellent performance by controlling the latent features or the attention maps of a denoising U-Net. Moreover, DMs can generate and edit images by adjusting latent features over multiple denoising steps~\cite{avrahami2022blended}. We focus on using latent features of DM, including intermediate features and cross-attention maps, across denoising steps to link them with the latent space of GAN and develop a multi-modal face image generation task.

\subsection{Multi-Modal Face Image Generation}
\vspace{-5pt}
Face generative models have progressed by incorporating various modalities, such as text~\cite{kim2022diffusionclip}, semantic mask~\cite{park2019semantic,wang2022semantic}, sketch~\cite{deng20233d,chen2020deepfacedrawing}, and audio~\cite{zhou2019talking}.
Several methods adopt StyleGAN, which can generate high-quality face images and edit facial attributes to control the style vectors. The transformer-based models~\cite{esser2021taming, bond2022unleashing} are also utilized, which improves the performance of face image generation by handling the correlation between multi-modal conditions using image quantization. A primary challenge faced in face generative models is to modify the facial attributes based on given conditions while minimizing changes to other attributes. Some methods~\cite{Patashnik_2021_ICCV, wei2022hairclip} edit facial attributes by manipulating the latent codes in GAN models. TediGAN~\cite{xia2021tedigan} controls multiple conditions by leveraging an encoder to convert an input image into latent codes and optimizing them with a pre-trained CLIP model. Recent works~\cite{nair2023unite,huang2023collaborative} use DMs to exploit the flexibility of taking multiple modalities as conditions and generate facial images directly from DMs.~Unlike existing methods, we use the pre-trained DM~\cite{zhang2023adding} as an encoder to further produce the latent codes for the pre-trained GAN models.

\section{Method}
\vspace{-2pt}
\subsection{Overview}
\vspace{-5pt}
Figure~\ref{fig:overview_model} illustrates the overall pipeline of our approach.
During the reverse diffusion process, we use the middle and decoder blocks of a denoising U-Net in ControlNet~\cite{zhang2023adding} as an encoder $\mathcal{E}$. A text prompt $c$, along with a visual condition $\mathbf{x}$, are taken as input to the denoising U-Net. Subsequently, $\mathcal{E}$ produces the feature maps $\mathbf{h}$ from the middle block, and the intermediate features $\mathbf{f}$ and the cross-attention maps $\mathbf{a}$ from the decoder blocks. $\mathbf{h}$ is then fed into the mapping network $\mathcal{M}$, which transforms the rich semantic feature into a latent code $\mathbf{w}^m$. The Attention-based Style Modulation Network (AbSMNet), $\mathcal{T}$, takes $\mathbf{f}$ and $\mathbf{a}$ as input to generate the modulation latent codes, $\mathbf{w^\gamma}$ and $\mathbf{w^\beta}$, that determine facial attributes related to the inputs. The latent code $\mathbf{w}$ is then forwarded to the pre-trained GAN $\mathcal{G}$ that generates the output image $I'$. Our model is trained across multiple denoising steps, and we use the denoising step $t$ to indicate the features and images obtained at each denoising step. With this pipeline, we aim to estimate the latent code, $\mathbf{w}^*_t$, that is used as input to $\mathcal{G}$ to render a GT image, $I^{gt}$:
\begin{equation}\label{equ:gan_inversion}
\mathbf{w}^*_t=\underset{\mathbf{w}_t}{\arg\min} \mathcal{L}(I^{gt}, \mathcal{G}(\mathbf{w}_t)),
\vspace{-6pt}
\end{equation}
where $\mathcal{L}(\cdot,\cdot)$ measures the distance between $I^{gt}$ and the rendered image, $I'= \mathcal{G}(\mathbf{w}_t)$. We employ learning-based \textit{GAN inversion} that estimates the latent code from an encoder to reconstruct an image according to given inputs.

\subsection{Mapping Network}
\label{sec:MappingNet}
\vspace{-5pt}
Our mapping network $\mathcal{M}$ aims to build a bridge between the latent space of the diffusion-based encoder $\mathcal{E}$ and that of the pre-trained GAN $\mathcal{G}$. $\mathcal{E}$ uses a text prompt and a visual input, and these textual and image embeddings are aligned by the cross-attention layers~\cite{zhang2023adding}. The feature maps $\mathbf{h}$ from the middle block of the denoising U-Net particularly contain rich semantics that resemble the latent space of the generator~\cite{kwon2022diffusion}. Here we establish the link between the latent spaces of $\mathcal{E}$ and $\mathcal{G}$ by using $\mathbf{h}_t$ across the denoising steps $t$.

Given $\mathbf{h}_t$, we design $\mathcal{M}$ that produces a 512-dimensional latent code $\mathbf{w}^m_t\in\mathbb{R}^{L\times512}$ that can be mapped to the latent space of $\mathcal{G}$:
\begin{equation}\label{equ:mapping}
\mathbf{w}^m_t=\mathcal{M}(\mathbf{h}_t).
\vspace{-6pt}
\end{equation}
$\mathcal{M}$ is designed based on the structure of the \textit{map2style} block in pSp~\cite{richardson2021encoding}, as seen in Figure~\ref{fig:overview_model}. This network consists of convolutional layers downsampling feature maps and a fully connected layer producing the latent code $\mathbf{w}^m_t$. 

\begin{figure}[!t]
    \centering    
    \includegraphics[width=0.96\linewidth]{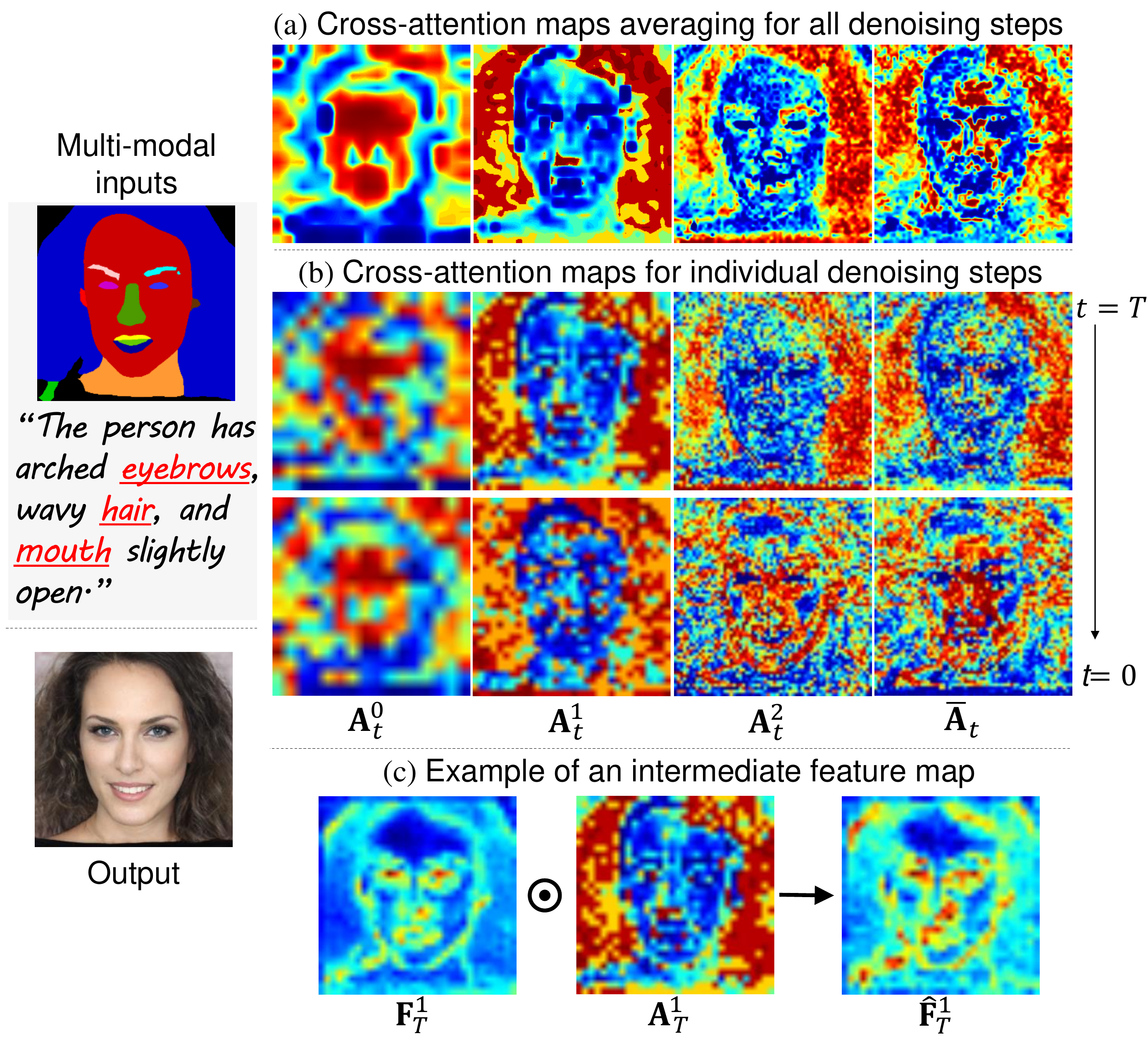}     
    \caption{Visualization of cross-attention maps and intermediate feature maps.
    (a) represents the semantic relation information between an input text and an input semantic mask in the spatial domain. The meaningful representations of inputs are shown across all denoising steps and $N$ different blocks. (b) represents $N$ different cross-attention maps, $\mathbf{A}_t$, at denoising steps $t=T$ and $t=0$. (c) shows the example of refined intermediate feature map $\hat{\mathbf{F}}^1_T$ at $1$st block and $t=T$ that is emphasized corresponding to input multi-modal conditions. The red and yellow regions of the map indicate higher attention scores. As the denoising step approaches $T$, the text-relevant features appear more clearly, and as the denoising step $t$ approaches 0, the features of the visual input are more preserved.}
    \vspace{-7pt}
    \label{fig:attention}
\end{figure}
\subsection{Attention-based Style Modulation Network}
\label{sec:AbSMNet}
\vspace{-5pt}
By training $\mathcal{M}$ with learning-based GAN inversion, we can obtain $\mathbf{w}^m_t$ and use it as input to the pre-trained GAN for image generation. However, we observe that $\mathbf{h}_t$ shows limitations in capturing fine details of the facial attributes due to its limited spatial resolution and data loss during the encoding. Conversely, the feature maps of the DM's decoder blocks show rich semantic representations~\cite{tumanyan2023plug}, benefiting from aggregating features from DM's encoder blocks via skip connections. We hence propose a novel Attention-based Style Modulation Network (AbSMNet), $\mathcal{T}$, that produces style modulation latent codes, $\mathbf{w}^\gamma_t, \mathbf{w}^\beta_t\in\mathbb{R}^{L\times512}$, by using $\mathbf{f}_t$ and $\mathbf{a}_t$ from $\mathcal{E}$. To improve reflecting the multi-modal representations to the final latent code $\mathbf{w}_t$, we modulate $\mathbf{w}^m_t$ from $\mathcal{M}$ using $\mathbf{w}^\gamma_t$ and $\mathbf{w}^\beta_t$, as shown in Figure~\ref{fig:overview_model}.

We extract intermediate features, $\mathbf{f}_t=\{{\mathbf{f}^n_t}\}^{N}_{n=1}$, from $N$ different blocks, and cross-attention maps, $\mathbf{a}_t=\{{\mathbf{a}^k_t}\}^{K}_{k=1}$, from $K$ different cross-attention layers of the $n$-th block, in $\mathcal{E}$ that is a decoder stage of denoising U-Net. The discriminative representations are represented more faithfully because $\mathbf{f}_t$ consists of $N$ multi-scale feature maps that can capture different sizes of facial attributes, which allows for finer control over face attributes. For simplicity, we upsample each intermediate feature map of $\mathbf{f}_t$ to same size intermediate feature maps $\mathbf{F}_t=\{{\mathbf{F}^n_t}\}^{N}_{n=1}$, where ${\mathbf{F}^n_t} \in\mathbb{R}^{H\times W\times C_n}$ has $H$, $W$, and $C_n$ as height, width and depth.

Moreover, $\mathbf{a}_t$ is used to amplify controlled facial attributes as it incorporates semantically related information in text and visual input.
To match the dimension with $\mathbf{F}_t$, we convert $\mathbf{a}_t$ to $\mathbf{A}_t=\{{\mathbf{A}^n_t}\}^{N}_{n=1}$, where $\mathbf{A}^n_t\in\mathbb{R}^{H\times W\times C_n}$, by max-pooling the output of the cross-attention layers in each decoder block and upsampling the max-pooling outputs. To capture the global representations, we additionally compute $\bar{\mathbf{A}}_t\in\mathbb{R}^{H\times W\times 1}$ by depth-wise averaging the max-pooling output of $\mathbf{a}_t$ over each word in the text prompt and upsampling it.
As illustrated in Figures~\ref{fig:attention}~(a) and (b), $\mathbf{A}_t$ and $\mathbf{\bar{A}}_t$ represent the specific regions aligned with input text prompt and visual input, such as semantic mask, across denoising steps $t$. By a pixel-wise multiplication between $\mathbf{F}_t$ and $\mathbf{A}_t$, we can obtain the refined intermediate feature maps $\hat{\mathbf{F}}_t$ that emphasize the representations related to multi-modal inputs as shown in Figure~\ref{fig:attention} (c). The improved average feature map $\hat{\bar{\mathbf{F}}}_t\in\mathbb{R}^{H\times W\times 1}$ is also obtained by multiplying $\mathbf{\bar{A}}_t$ with $\mathbf{\bar{F}}_t$, where $\mathbf{\bar{F}}_t\in\mathbb{R}^{H\times W\times 1}$ is obtained by first averaging the feature maps in $\mathbf{F}_t=\{{\mathbf{F}^n_t}\}^{N}_{n=1}$ and then depth-wise averaging the outputs. $\hat{\mathbf{F}}_t$ and $\hat{\bar{\mathbf{F}}}_t$ distinguish text- and structural-relevant semantic features, which improves the alignment with the inputs.
\begin{figure}[!t]
    \centering    
    \includegraphics[width=0.95\linewidth]{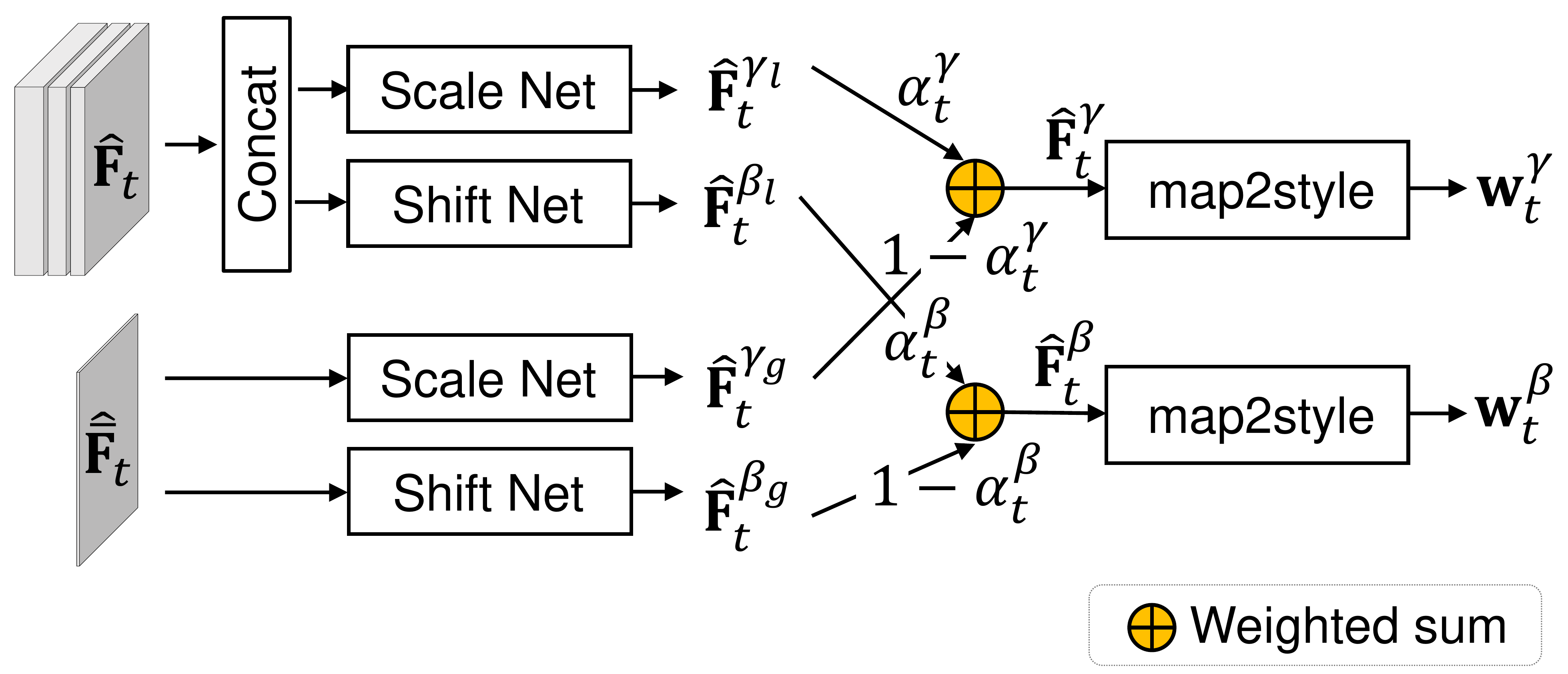} 
    \caption{Style modulation network in $\mathcal{T}$. The refined intermediate feature maps $\hat{\mathbf{F}}_t$ and $\hat{\bar{\mathbf{F}}}_t$ are used to capture local and global semantic representations, respectively. They are fed into the scale and shift network, respectively. The weighted summations of these outputs are used as input to the map2style network, which finally generates the scale and shift modulation latent codes, $\mathbf{w}^\gamma_t$, and $\mathbf{w}^\beta_t$.}
    \vspace{-6pt}
    \label{fig:stylenet}
\end{figure}
~We use $\hat{\mathbf{F}}_t$ and $\hat{\bar{\mathbf{F}}}_t$ as input to the style modulation network that produces the modulation codes $\mathbf{w}^\gamma_t$, and $\mathbf{w}^\beta_t$ as shown in Figure~\ref{fig:stylenet}.
We capture both local and global features by using $\hat{\mathbf{F}}_t$, which consists of feature maps representing different local regions on the face, and $\hat{\bar{\mathbf{F}}}_t$, which implies representations of the entire face. We concatenate $N$ intermediate feature maps of $\hat{\mathbf{F}}_t$, $concat(\hat{\mathbf{F}}^1_t \cdots \hat{\mathbf{F}}^{N}_t)$, and it is forward to the scale and shift networks that consist of convolutional layers and Leaky ReLU, forming the local modulation feature maps, $\hat{\mathbf{F}}^{\gamma_l}_t$ and $\hat{\mathbf{F}}^{\beta_l}_t$. We also estimate global modulation feature maps, $\hat{\mathbf{F}}^{\gamma_g}_t$ and $\hat{\mathbf{F}}^{\beta_g}_t$, by feeding $\hat{\bar{\mathbf{F}}}_t$ to the scale and shift network. The final scale, $\hat{\mathbf{F}}^{\gamma}_t$, and shift, $\hat{\mathbf{F}}^{\beta}_t$, feature maps are estimated by the weighted summation:
\begin{flalign}\label{equ:modulation_feature} 
{}&\hat{\mathbf{F}}^\gamma_t=\alpha^\gamma_t\hat{\mathbf{F}}^{\gamma_l}_t+(1-\alpha^\gamma_t)\hat{\mathbf{F}}^{\gamma_g}_t,\\ \nonumber 
{}&\hat{\mathbf{F}}^\beta_t=\alpha^\beta_t\hat{\mathbf{F}}^{\beta_g}_t+(1-\alpha^\beta_t)\hat{\mathbf{F}}^{\beta_g}_t,
\end{flalign}
where $\alpha^\gamma_t$ and $\alpha^\beta_t$ are learnable weight parameters.
Through the \textit{map2style} module, we then convert $\hat{\mathbf{F}}^\gamma_t$ and $\hat{\mathbf{F}}^\beta_t$ into the final scale, $\mathbf{w}^\gamma_t\in\mathbb{R}^{L\times512}$, and shift, $\mathbf{w}^\beta_t\in\mathbb{R}^{L\times512}$, latent codes.
With these modulation latent codes, we achieve more precise control over facial details while corresponding to the input multi-modal inputs at the pixel level.

Finally, the mapped latent code $\mathbf{w}^m_t$ from $\mathcal{M}$ is modulated by $\mathbf{w}^\gamma_t$ and $\mathbf{w}^\beta_t$ from $\mathcal{T}$ to get the final latent code $\mathbf{w}_t$ that is used to obtain the generated image $I'_t$ as follows:
\begin{flalign}\label{equ:modulation} 
{}&\mathbf{w}_t= \mathbf{w}^m_t\odot\mathbf{w}^\gamma_t\oplus  \mathbf{w}^\beta_t,\\
{}& {I'_t}=\mathcal{G}(\mathbf{w}_t).  
\end{flalign}

\begin{figure*}[!hbt]
    \centering
    \includegraphics[width=0.89\linewidth]{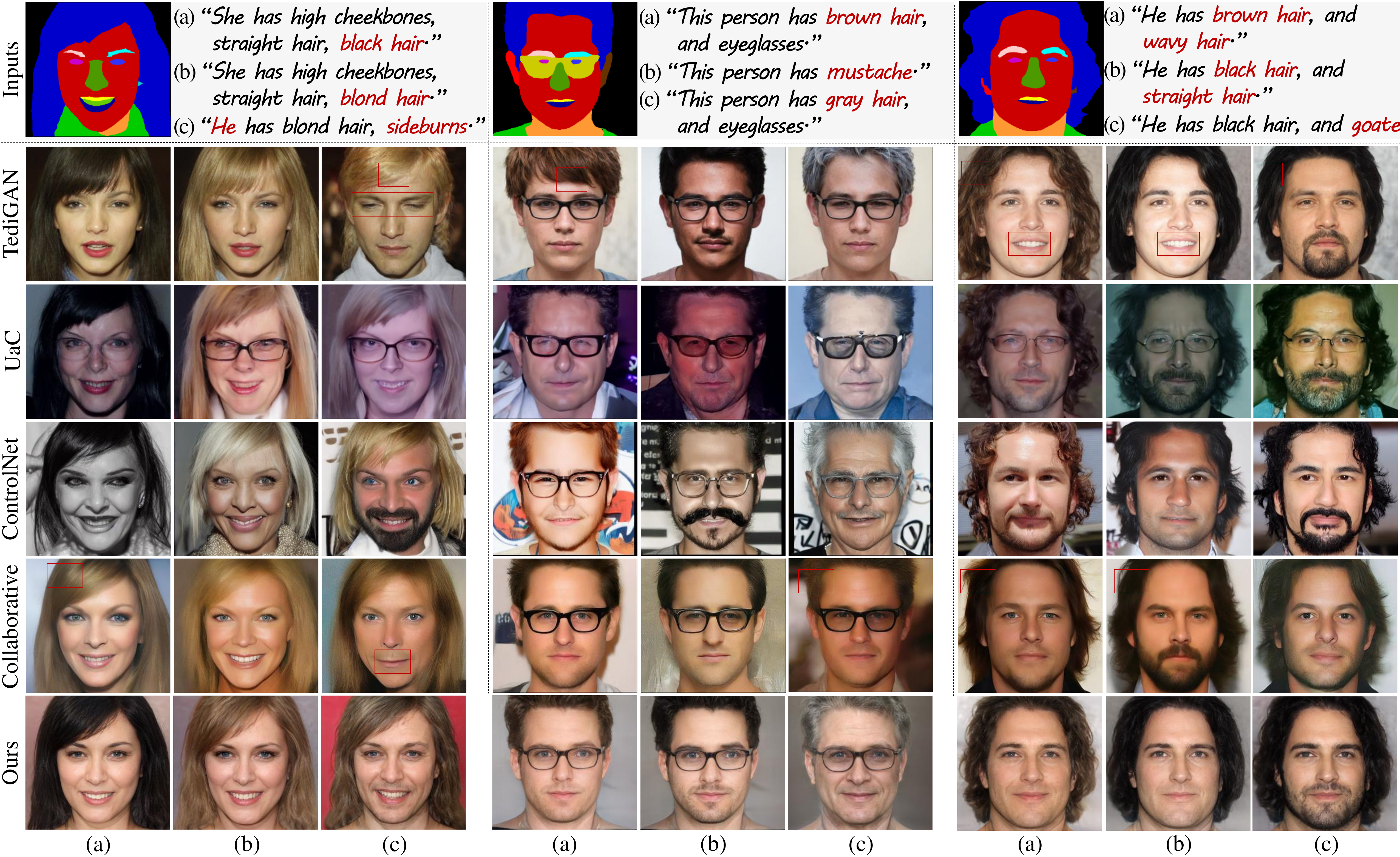}
    \vspace{-7pt}
    \caption{Visual examples of the 2D face image generation using a text prompt and a semantic mask. For each semantic mask, we use three different text prompts (a)-(c), resulting in different output images (a)-(c).}    
    \label{fig:result_local_editing}
    \vspace{-7pt}
\end{figure*}
\subsection{Loss Functions}
\label{sec:Losses}
\vspace{-3pt}
To optimize $\mathcal{M}$ and $\mathcal{T}$, we use reconstruction loss, perceptual loss, and identity loss for image generation, and regularization loss~\cite{richardson2021encoding} that encourages the latent codes to be closer to the average latent code $\mathbf{\bar{w}}$.

For training $\mathcal{M}$, we use the GT image $I^{gt}$ as reference to encourage the latent code $\mathbf{w}^m_t$ to generate a photo-realistic image as follows:
\begin{flalign}\label{equ:mapping_loss_m} 
\mathcal{L}_{\mathcal{M}}= {}&\lambda^m_0\Vert{I^{gt}- \mathcal{G}(\mathbf{w}^m_t})\Vert_2+\\ \nonumber {}&\lambda^m_1\Vert{\mathcal{F}(I^{gt})-\mathcal{F}(\mathcal{G}(\mathbf{w}^m_t)}\Vert_2+\\ \nonumber
{}&\lambda^m_2(1-cos(\mathcal{R}(I^{gt}), \mathcal{R}(\mathcal{G}(\mathbf{w}^m_t))))+\\ \nonumber
{}&\lambda^m_3\Vert\mathcal{E}(\mathbf{z}_t,t,\mathbf{x},c)-\mathbf{\bar{w}}\Vert_2,
\end{flalign}
where $\mathcal{R}(\cdot)$ is pre-trained ArcFace network~\cite{deng2019arcface}, $\mathcal{F}(\cdot)$ is the feature extraction network~\cite{zhang2018unreasonable}, $z_t$ is noisy image, and the hyper-parameters $\lambda^m_{(\cdot)}$ guide the effect of losses. Note that we freeze $\mathcal{T}$ while training $\mathcal{M}$.

For training $\mathcal{T}$, we use $I^{d}_{0}$ produced by the encoder $\mathcal{E}$ into the reconstruction and perceptual losses. With these losses, the loss $\mathcal{L}_{\mathcal{T}}$ encourages the network to control facial attributes while preserving the identity of $I^{gt}$: 
\begin{flalign}\label{equ:mapping_loss_t} 
\mathcal{L}_{\mathcal{T}}= {}&\lambda^s_0\Vert{I^{d}_{0}- \mathcal{G}(\mathbf{w}_t})\Vert_2+\\ \nonumber {}&\lambda^s_1\Vert{\mathcal{F}(I^{d}_{0})-\mathcal{F}(\mathcal{G}(\mathbf{w}_t)}\Vert_2+\\ \nonumber
{}&\lambda^s_2(1-cos(\mathcal{R}(I^{gt}), \mathcal{R}(\mathcal{G}(\mathbf{w}_t))))+\\ \nonumber
{}&\lambda^s_3\Vert\mathcal{E}(\mathbf{z}_t,t,\mathbf{x},c)-\mathbf{\bar{w}}\Vert_2,
\end{flalign}
where the hyper-parameters $\lambda^s_{(\cdot)}$ guide the effect of losses. Similar to Equation~\ref{equ:mapping_loss_m}, we freeze $\mathcal{M}$ while training $\mathcal{T}$.

We further introduce a multi-step training strategy that considers the evolution of the feature representation in $\mathcal{E}$ over the denoising steps. We observe that $\mathcal{E}$ tends to focus more on text-relevant features in an early step, $t=T$, and structure-relevant features in a later step, $t=0$. Figure~\ref{fig:attention}~(b) shows the attention maps $\bar{\mathbf{A}}$ showing variations across the denoising step. As the attention map, we can capture the textual and structural features by varying the denoising steps.
To effectively capture the semantic details of multi-modal conditions, our model is trained across multiple denoising steps.

\section{Experiments}
\vspace{-2pt}
\subsection{Experimental Setup}
\vspace{-5pt}
We use ControlNet~\cite{zhang2023adding} as the diffusion-based encoder that receives multi-modal conditions, including text and visual conditions such as a semantic mask and scribble map. The StyleGAN~\cite{karras2020analyzing} and EG3D~\cite{chan2022efficient} are exploited as pre-trained 2D and 3D GAN, respectively. See the Supplementary Material for the training details, the network architecture, and additional results.

\begin{figure*}[!hbt]
    \centering 
    \includegraphics[width=0.93\linewidth]{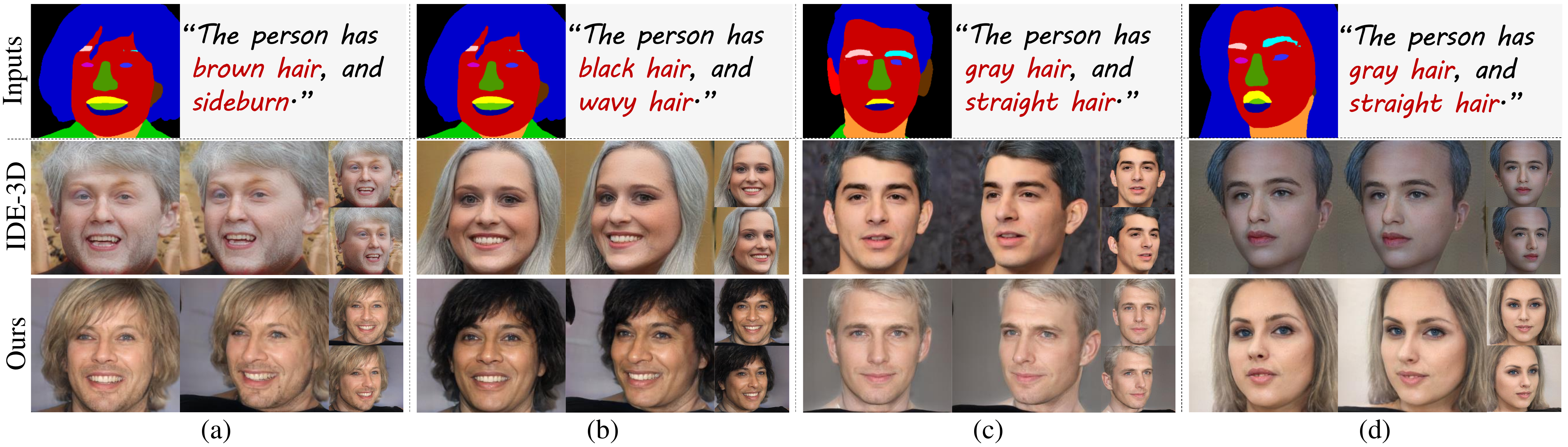} 
      \vspace{-5pt}
    \caption{Visual examples of the 3D-aware face image generation using a text and a semantic mask. We show the images generated with inputs and arbitrary viewpoints.}
    \label{fig:result_local_editing_3d}   
\end{figure*}
\begin{table*}[!hbt]
\begin{center}  
\small
\begin{tabular}{cllccccccc}
 \hlinewd{1pt}
\multicolumn{1}{c}{Input conditions}                        & Method    & Model &Domain& FID$\downarrow$& LPIPS$\downarrow$ & SSIM$\uparrow$ & ID$\uparrow$ &ACC$\uparrow$& mIoU$\uparrow$\\ \cline{1-10} 
{\multirow{7}{*}{\makecell{Text + \\ semantic mask }}} & TediGAN~\cite{xia2021tedigan}& GAN &2D   & 54.83 & 0.31 & 0.62& 0.63 &81.68 & 40.01  \\
& IDE-3D~\cite{sun2022ide}& GAN &3D   & \textbf{39.05} & 0.40 & 0.41 & 0.54 & 47.07 & 10.98  \\
& UaC~\cite{nair2023unite} & Diffusion&2D   & 45.87 & 0.38  & 0.59 &0.32 &81.49 & 42.68  \\
& ControlNet~\cite{zhang2023adding} & Diffusion&2D   & 46.41 & 0.41 & 0.53 &0.30 &82.42&42.77  \\
& Collaborative~\cite{huang2023collaborative} & Diffusion&2D   &  48.23 & 0.39 & 0.62 &0.31  & 74.06 &30.69  \\
\rowcolor{mylightblue} \cellcolor{white}& Ours  & GAN &2D   & 46.68 & \underline{0.30}  & \underline{0.63} & \underline{0.76}&\textbf{83.41} &\textbf{43.82}  \\
\rowcolor{mylightblue} \cellcolor{white} & Ours & GAN &3D & \underline{44.91} & \textbf{0.28}  & \textbf{0.64}& \textbf{0.78}& \underline{83.05} & \underline{43.74}  \\
\hlinewd{1pt}
{\multirow{3}{*}{\makecell{Text + \\ scribble map}}} & ControlNet~\cite{zhang2023adding}  & Diffusion & 2D&   93.26  &  0.52   & 0.25 &0.21 &- &  -  \\
& \cellcolor{mylightblue}Ours      & \cellcolor{mylightblue}GAN & \cellcolor{mylightblue}2D& \cellcolor{mylightblue}\underline{55.60}  &  \cellcolor{mylightblue}\textbf{0.32}  &  \cellcolor{mylightblue}\textbf{0.56}  & \cellcolor{mylightblue}\textbf{0.72} & \cellcolor{mylightblue}- &\cellcolor{mylightblue}-    \\
& \cellcolor{mylightblue}Ours  & \cellcolor{mylightblue}GAN & \cellcolor{mylightblue}3D& \cellcolor{mylightblue}\textbf{48.76}  &  \cellcolor{mylightblue}\underline{0.34}  &  \cellcolor{mylightblue}\underline{0.49}  & \cellcolor{mylightblue}\underline{0.62} & \cellcolor{mylightblue}- &\cellcolor{mylightblue}-    \\
 \hlinewd{1pt}
\end{tabular}
 \end{center}   
 \vspace{-13pt}
  \caption{Quantitative results of multi-modal face image generation on CelebAMask-HQ~\cite{CelebAMask-HQ} with annotated text prompts~\cite{xia2021tedigan}. } 
    \label{table:all_method}
    \vspace{-7pt}
\end{table*}
\noindent\textbf{Datasets.}
We employ the CelebAMask-HQ~\cite{CelebAMask-HQ} dataset comprising 30,000 face RGB images and annotated semantic masks, including 19 facial-component categories such as \textit{skin}, \textit{eyes}, \textit{mouth}, and \textit{etc}. 
We also use textual descriptions provided by~\cite{xia2021tedigan} describing the facial attributes, such as \textit{black hair}, \textit{sideburns}, and \textit{etc}, corresponding to the CelebAMask-HQ dataset.
For the face image generation task using a scribble map, we obtain the scribble maps by applying PiDiNet~\cite{su2021pdc, su2019bird} to the RGB images in CelebAMask-HQ.
We additionally compute camera parameters based on~\cite{deng2019accurate, chan2022efficient} for 3D-aware image generation.

\noindent\textbf{Comparisons.}
We compare our method with GAN-based models, such as TediGAN~\cite{xia2021tedigan} and IDE-3D~\cite{sun2022ide}, and DM-based models, such as Unite and Conquer (UaC)~\cite{nair2023unite}, ControlNet~\cite{zhang2023adding}, and Collaborative diffusion (Collaborative)~\cite{huang2023collaborative},  for face generation task using a semantic mask and a text prompt. IDE-3D is trained by a CLIP loss term like TediGAN to apply a text prompt for 3D-aware face image generation. ControlNet is used for face image generation using a text prompt and a scribble map. We use the official codes provided by the authors, and we downsample the results into $256\times 256$ for comparison.

\noindent\textbf{Evaluation Metrics.}\label{sec:exp_eval}
For quantitative comparisons, we evaluate the image quality and semantic consistency using sampled 2k semantic mask- and scribble map-text prompt pairs.
Frechet Inception Distance (FID)~\cite{heusel2017gans}, LPIPS~\cite{zhang2018unreasonable}, and the Multiscale Structural Similarity (MS-SSIM)~\cite{wang2003multiscale} are employed for the evaluation of visual quality and diversity, respectively. 
We also compute the ID similarity mean score (ID)~\cite{deng2019arcface, wei2022hairclip} before and after applying a text prompt.
Additionally, we assess the alignment accuracy between the input semantic masks and results using mean Intersection-over-Union (mIoU) and pixel accuracy (ACC) for the face generation task using a semantic mask.

\subsection{Results}\label{sec:experimentalresults}
\vspace{-5pt}
\noindent\textbf{Qualitative Evaluations.}
Figure~\ref{fig:result_local_editing} shows the visual comparisons between ours and two existing methods for 2D face image generation using a text prompt and a semantic mask as input. We use the same semantic mask with different text prompts (a)-(c). TediGAN produces results consistent with the text prompt as the latent codes are optimized using the input text prompt. However, the results are inconsistent with the input semantic mask, as highlighted in the red boxes. UaC shows good facial alignment with the input semantic mask, but the results are generated with unexpected attributes, such as glasses, that are not indicated in the inputs. Collaborative and ControlNet produce inconsistent, blurry, and unrealistic images. Our model is capable of preserving semantic consistency with inputs and generating realistic facial images. As shown in Figure~\ref{fig:result_local_editing}, our method preserves the structure of the semantic mask, such as the hairline, face position, and mouth shape, while changing the attributes through a text prompt.

Figure~\ref{fig:result_local_editing_3d} compares our method with IDE-3D~\cite{sun2022ide} to validate the performance of 3D-aware face image generation using a semantic mask and a text prompt. We use the same semantic mask with different text prompts in Figures~\ref{fig:result_local_editing_3d}~(a)~and~(b), and use the same text prompt with different semantic masks in Figures~\ref{fig:result_local_editing_3d}~(c) and (d). The results of IDE-3D are well aligned with the semantic mask with the frontal face. However, IDE-3D fails to produce accurate results when the non-frontal face mask is used as input. Moreover, the results cannot reflect the text prompt. Our method can capture the details provided by input text prompts and semantic masks, even in a 3D domain. 

\begin{figure}[!t]
    \centering
    \includegraphics[width=0.95\linewidth]{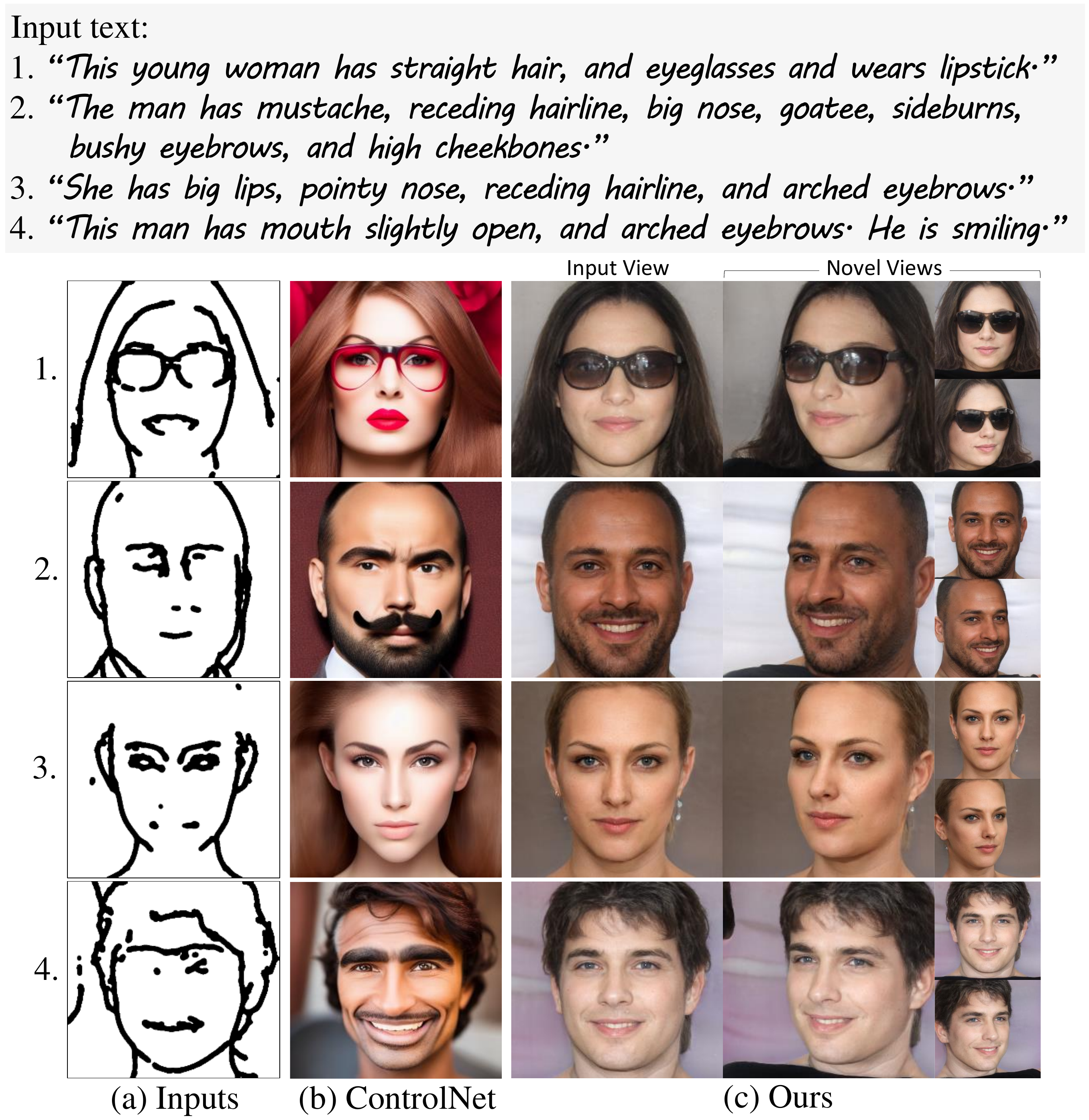}
    \vspace{-3pt}
       \caption{Visual examples of 3D-aware face image generation using text prompts and scribble maps. Using (1-4) the text prompts and their corresponding (a) scribble maps, we compare the results of (b) ControlNet with (c) multi-view images generated by ours.}
    \label{fig:result_scribble}
    \vspace{-10pt}
\end{figure}

Figure~\ref{fig:result_scribble} shows visual comparisons with ControlNet on 2D face generation from a text prompt and a scribble map. The results from ControlNet and our method are consistent with both the text prompt and the scribble map. ControlNet, however, tends to over-emphasize the characteristic details related to input conditions. Our method can easily adapt to the pre-trained 3D GAN and produce photo-realistic multi-view images from various viewpoints.

\noindent\textbf{Quantitative Evaluations.}
Table~\ref{table:all_method} reports the quantitative results on CelebAMask-HQ with text prompts~\cite{xia2021tedigan}. Our method using text prompts and semantic masks shows performance increases in all metrics in 2D and 3D domains, compared with TediGAN and UaC. Our model using 2D GAN significantly improves LPIPS, ID, ACC, and mIoU scores, surpassing TediGAN, UaC, ControlNet, and Collaborative, respectively. It demonstrates our method's strong ability to generate photo-realistic images while reflecting input multi-modal conditions better. For 3D-aware face image generation using a text prompt and a semantic mask, it is reasonable that IDE-3D shows the highest FID score as the method additionally uses an RGB image as input to estimate the latent code for face generation. The LPIPS, SSIM, and ID scores are significantly higher than IDE-3D, with scores higher by 0.116, 0.23, and 0.24, respectively. Our method using 3D GAN exhibits superior ACC and mIoU scores for the 3D face generation task compared to IDE-3D, with the score difference of 35.98$\%$ and 32.76$\%$, likely due to its ability to reflect textual representations into spatial information. In face image generation tasks using a text prompt and a scribble map, our method outperforms ControlNet in FID, LPIPS, SSIM, and ID scores in both 2D and 3D domains. Note that the ACC and mIoU scores are applicable for semantic mask-based methods.

\begin{figure}[!t]
    \centering    
    \begin{minipage}{0.5\textwidth}
    \includegraphics[width=0.92\linewidth]{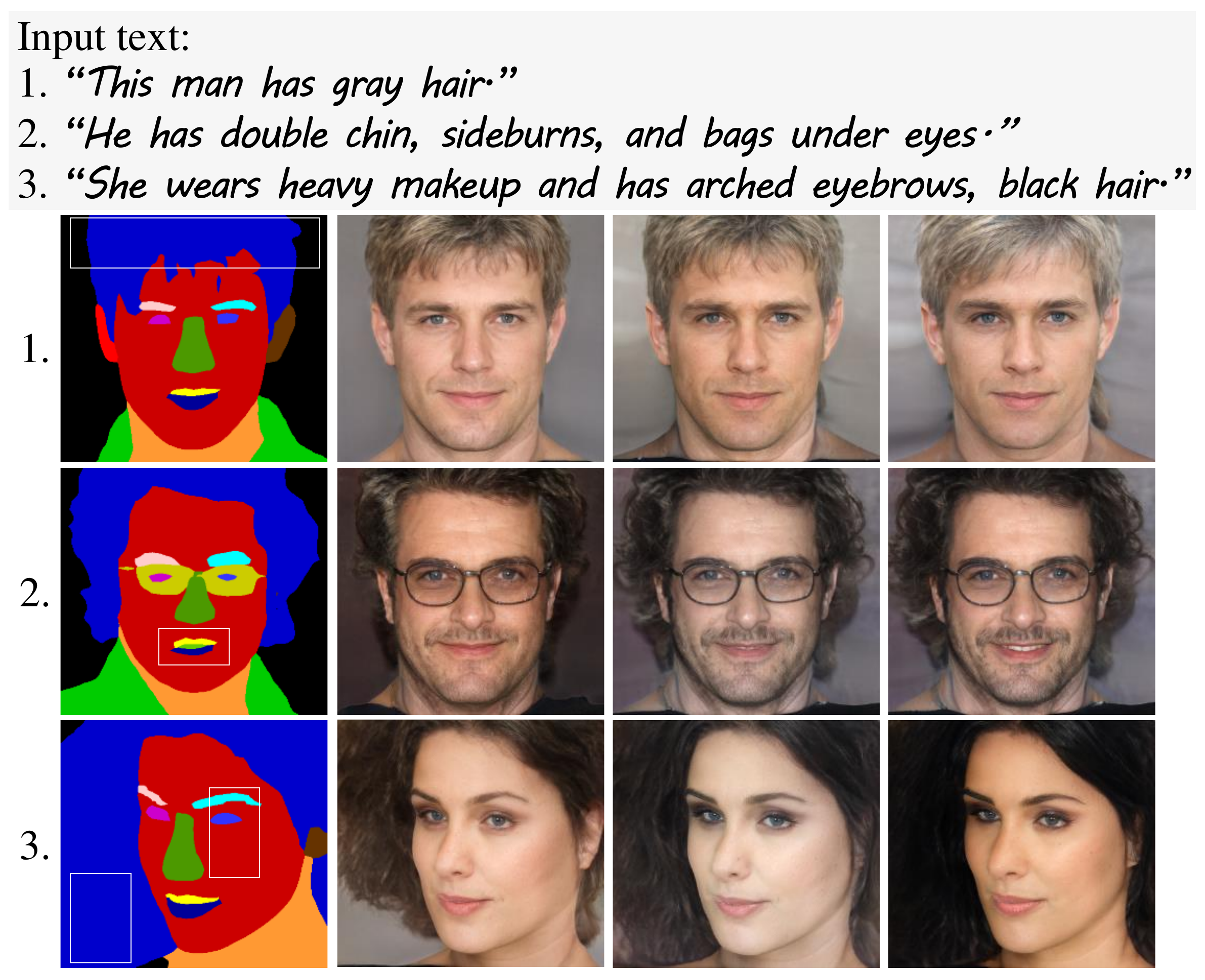} 
   \end{minipage}\vfill
   \begin{minipage}{0.5\textwidth}
   \begin{minipage}{0.25\textwidth}
   \centering\small{\quad(a) Inputs}
   \end{minipage}\hfill
   \begin{minipage}{0.25\textwidth}
   \small{\quad(b) w/o $\mathcal{T}$}
   \end{minipage}\hfill\hspace{-5pt}
   \begin{minipage}{0.25\textwidth}
   \small{(c) w/o $\mathbf{A}$, $\bar{\mathbf{A}}$}
   \end{minipage}\hfill
   \begin{minipage}{0.25\textwidth}
   \small{(d) Ours}
   \end{minipage}\hfill
   \end{minipage}\vfill
   \vspace{-2pt}
    \caption{Effect of $\mathcal{M}$ and $\mathcal{T}$. (b) shows the results using only $\mathcal{M}$, and (c) shows the effect of the cross-attention maps ($\mathbf{A}$ and $\bar{\mathbf{A}}$) in $\mathcal{T}$. The major changes are highlighted with the white boxes.}  
    \label{fig:result_abl}
\end{figure}
\begin{table}[!t]
\begin{center} 
\small
\begin{tabular}{M{0.9cm} | M{0.1cm}M{0.1cm}M{0.1cm}M{0.1cm}M{0.15cm}|M{0.6cm}M{0.8cm}M{0.5cm}M{0.6cm}}
\hlinewd{1pt}
$\textrm{Method}$ &$\mathcal{M}$ & $\mathcal{T}$ & $\mathbf{A}_t$ & ${I}^{gt}$ &${I}^d_{0}$ & FID$\downarrow$&   \centering{LPIPS$\downarrow$} & ID$\uparrow$ & ACC$\uparrow$ \\ \cline{1-10}  
(a)  & \checkmark &  &  &  \checkmark&  \checkmark&62.08&\centering{0.29} & 0.62 & 81.09   \\
(b)& \checkmark & \checkmark &  & \checkmark  &\checkmark&48.68&\centering{0.28} &  0.66 & 82.86  \\
\cline{1-10} 
\rowcolor{white} (c)& \checkmark & \checkmark&  \checkmark & & \checkmark &  54.27 &\centering{0.31}& 0.58 &80.58  \\
(d)&\checkmark & \checkmark&  \checkmark & \checkmark&  & 61.60& \centering{0.29}& 0.62 & 80.04  \\
\cline{1-10} 
\rowcolor{mylightblue}  (e)& \checkmark & \checkmark &  \checkmark & \checkmark & \checkmark &$\textbf{44.91}$& \centering{$\textbf{0.28}$}  & \textbf{0.78} & $\textbf{83.05}$ \\
\hlinewd{1pt}
\end{tabular}
\end{center}  
\vspace{-15pt}
\caption{Ablation analysis on 3D-aware face image generation using a text prompt and a semantic mask. We compare (a) and (b) with (e) to show the effect of our style modulation network and (c) and (d) with (e) to analyze the effect of $I^{gt}$ and $I^d$ in model training.}  
\vspace{-5pt}
\label{table:ablation}
\end{table}
\subsection{Ablation Study}
\vspace{-3pt}
We conduct ablation studies to validate the effectiveness of our contributions, including the mapping network $\mathcal{M}$, the AbSM network $\mathcal{T}$, and the loss functions $\mathcal{L}_{\mathcal{M}}$ and $\mathcal{L}_{\mathcal{T}}$.

\noindent\textbf{Effectiveness of $\mathcal{M}$ and $\mathcal{T}$.}
We conduct experiments with different settings to assess the effectiveness of $\mathcal{M}$ and $\mathcal{T}$. We also show the advantages of using cross-attention maps in our model. The quantitative and qualitative results are presented in Table~\ref{table:ablation} and Figure~\ref{fig:result_abl}, respectively.
When using only $\mathcal{M}$, we can generate face images that roughly preserve the structures of a given semantic mask in Figure~\ref{fig:result_abl}~(a), including the outline of the facial components ($\eg$ face, eye) in Figure~\ref{fig:result_abl}~(b). On the other hand, $\mathcal{T}$ enables the model to express face attribute details effectively, such as hair colors and mouth open, based on the multi-modal inputs in Figure~\ref{fig:result_abl}~(c). The FID and ACC scores are higher than the model using only $\mathcal{M}$ in Table~\ref{table:ablation} (b). We further present the impact of adopting cross-attention maps to $\mathcal{T}$ for style modulation. Figure~\ref{fig:result_abl}~(d) shows how the attention-based modulation approach enhances the quality of results, particularly in terms of the sharpness of desired face attributes and the overall consistency between the generated image and multi-modal conditions. Table~\ref{table:ablation} (e) demonstrates the effectiveness of our method by showing improvements in FID, LPIPS, ID, and ACC. Our method, including both $\mathcal{M}$ and $\mathcal{T}$ with cross-attention maps, significantly improves the FID showing our model's ability to generate high-fidelity images. 
From the improvement of the ID score, the cross-attention maps enable relevantly applying the details of input conditions to facial components.

\begin{figure}[!t]
    \centering    
    \includegraphics[width=0.93\linewidth]{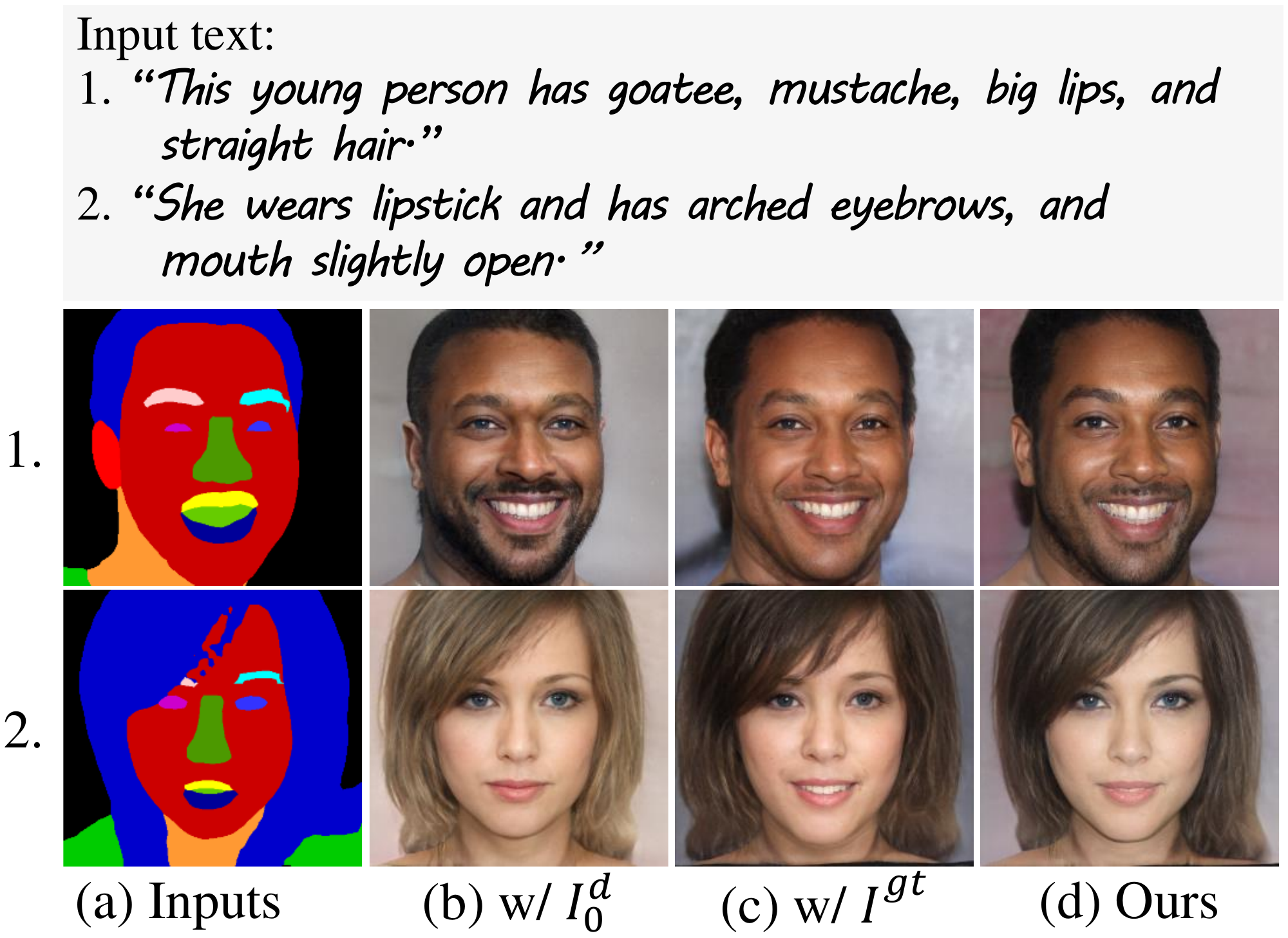} 
    \vspace{-5pt}
   \caption{Effect of using $I^{d}$ from the denoising U-Net and the GT image $I^{gt}$ in model training. Using text prompts (1, 2) with (a) the semantic mask, we show face images using our model trained with (b) $I^{d}_0$, (c) $I^{gt}$, and (d) both.}
   \vspace{-10pt}
    \label{fig:result_abl_loss}
\end{figure}
\noindent\textbf{Model Training.} 
We analyze the effect of loss terms $\mathcal{L}_{\mathcal{M}}$ and $\mathcal{L}_{\mathcal{T}}$ by comparing the performance with the model trained using either $I^d_{0}$ from the denoising U-Net or GT image $I^{gt}$. The model trained using $I^d_{0}$ produces the images in Figure~\ref{fig:result_abl_loss} (b), which more closely reflected the multi-modal conditions (a), such as ``goatee'' and ``hair contour''. In Table~\ref{table:ablation} (c), the ACC score of this model is higher than the model trained only using $I^{gt}$ in Table~\ref{table:ablation}~(d).
The images generated by the model trained with $I^{gt}$ in Figure~\ref{fig:result_abl_loss}~(c) are more perceptually realistic, as evidenced by the lower LPIPS score compared to the model trained with $I^d_{0}$ in Table~\ref{table:ablation}~(c) and (d). Using $I^{gt}$ also preserves more condition-irrelevant features inferred by the ID scores in Table~\ref{table:ablation}~(c) and (d). In particular, our method combines the strengths of two models as shown in Figure~\ref{fig:result_abl_loss} (d) and Table~\ref{table:ablation} (e).

\subsection{Limitations and Future Works}
\begin{figure}[!t]
    \centering    
    \includegraphics[width=0.89\linewidth]{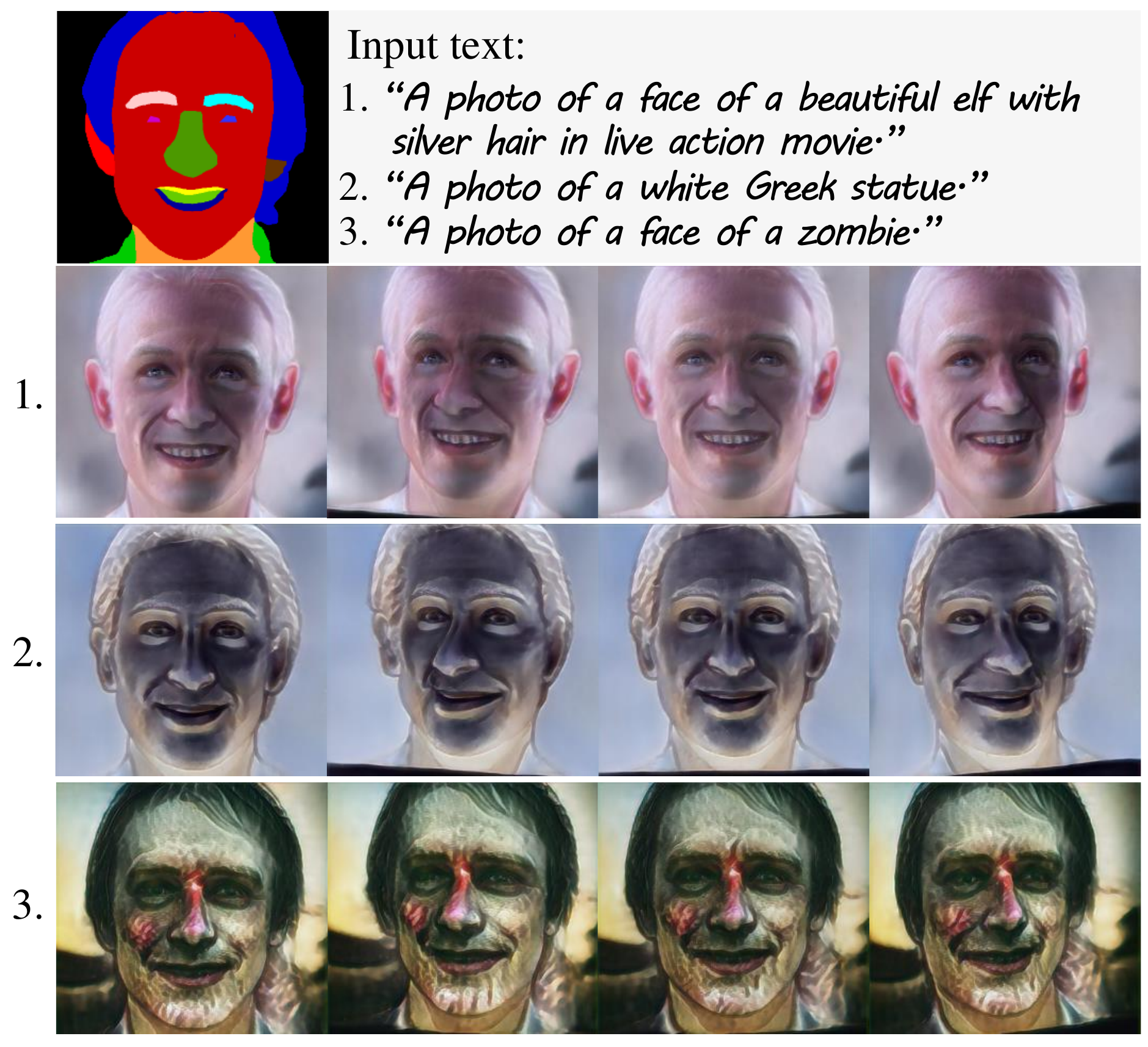} 
\caption{Visual examples of 3D face style transfer. Our method generates stylized multi-view images by mapping the latent features of DM and GAN.}
\vspace{-5pt}
    \label{fig:result_future_style}
\end{figure}

\vspace{-5pt}
Our method can be extended to multi-modal face style transfer ($\eg$ \textit{face} $\rightarrow$ \textit{Greek statue}) by mapping the latent spaces of DM and GAN without CLIP losses and additional dataset, as shown in Figure~\ref{fig:result_future_style}. For the 3D-aware face style transfer task, we train our model using $I^d_{0}$ that replaces GT image $I^{gt}$ in our loss terms. This method, however, is limited as it cannot transfer extremely distinct style attributes from the artistic domain to the photo-realistic domain of GAN. To better transfer the facial style in the 3D domain, we will investigate methods to map the diffusion features related to the input pose into the latent space of GAN in future works.
 
\section{Conclusion}
\vspace{-5pt}
We presented the diffusion-driven GAN inversion method that translates multi-modal inputs into photo-realistic face images in 2D and 3D domains. Our method interprets the pre-trained GAN's latent space and maps the diffusion features into this latent space, which enables the model to easily adopt multi-modal inputs, such as a visual input and a text prompt, for face image generation. We also proposed to train our model across the multiple denoising steps, which further improves the output quality and consistency with the multiple inputs. We demonstrated the capability of our method by using text prompts with semantic masks or scribble maps as input for 2D or 3D-aware face image generation and style transfer. 

{
    \small
    \bibliographystyle{ieeenat_fullname}
    \bibliography{main}
}

\end{document}